\documentclass[10pt,twocolumn,letterpaper]{article}

\usepackage{iccv}
\usepackage{times}
\usepackage{epsfig}
\usepackage{graphicx}
\usepackage{amsmath}
\usepackage{amssymb}

\usepackage{enumitem}
\usepackage{subcaption}
\usepackage{adjustbox}
\setlist[enumerate,1]{label=\arabic*}
\newlist{inlinelist}{enumerate*}{1}
\setlist*[inlinelist,1]{label=(\roman*)}

\usepackage{diagbox}
\usepackage{multirow}

\newcommand{\modelname}{TASED-Net}
\newcommand{\modelnamelong}{Temporally-Aggregating Spatial Encoder-Decoder}
\newcommand{\auxpnamet}{Auxiliary pooling}
\newcommand{\auxpname}{\textit{\auxpnamet}}
\mathchardef\mhyphen="2D
\usepackage[breaklinks=true,bookmarks=false]{hyperref}

\iccvfinalcopy 

\def\iccvPaperID{2157} 

\ificcvfinal\fi

\begin{document}

\title{\modelname{}: \modelnamelong{} Network \\for Video Saliency Detection}

\author{Kyle Min\quad Jason J. Corso\\
University of Michigan\\
Ann Arbor, MI 48109\\
{\tt\small \{kylemin,jjcorso\}@umich.edu}
}

\maketitle
\ificcvfinal\thispagestyle{empty}\fi

\begin{abstract}
   \modelname{} is a 3D fully-convolutional network architecture for video saliency detection. It consists of two building blocks: first, the encoder network extracts low-resolution spatiotemporal features from an input clip of several consecutive frames, and then the following prediction network decodes the encoded features spatially while aggregating all the temporal information. As a result, a single prediction map is produced from an input clip of multiple frames. Frame-wise saliency maps can be predicted by applying \modelname{} in a sliding-window fashion to a video. The proposed approach assumes that the saliency map of any frame can be predicted by considering a limited number of past frames. The results of our extensive experiments on video saliency detection validate this assumption and demonstrate that our fully-convolutional model with temporal aggregation method is effective. \modelname{} significantly outperforms previous state-of-the-art approaches on all three major large-scale datasets of video saliency detection: DHF1K, Hollywood2, and UCFSports. After analyzing the results qualitatively, we observe that our model is especially better at attending to salient moving objects.
\end{abstract}

\section{Introduction} \label{sec:intro}
Video saliency detection aims to model the gaze fixation patterns of humans when viewing a dynamic scene. Because the predicted saliency map can be used to prioritize the video information across space and time, this task has a number of applications such as video surveillance~\cite{guraya2010predictive, yubing2011spatiotemporal}, video captioning~\cite{nguyen2013static}, video compression~\cite{guo2010novel, hadizadeh2014saliency}, etc.

Previous state-of-the-art approaches for video saliency detection~\cite{bazzani2016recurrent, jiang2017predicting, wang2018revisiting} largely depend on LSTMs~\cite{hochreiter1997long} to aggregate information temporally. For example, OM-CNN~\cite{jiang2017predicting} feeds spatial features from YOLO~\cite{redmon2016you} and temporal features from FlowNet~\cite{dosovitskiy2015flownet} into a two-layer LSTM. The leading state-of-the-art model, ACLNet~\cite{wang2018revisiting}, also uses a LSTM to aggregate spatial features guided by frame-wise image saliency maps. The strong performance of LSTM-based approaches over non-LSTM based ones suggests that aggregating information temporally boosts performance on video saliency detection.

However, all of these LSTM-based, existing video saliency models fail to jointly process spatial and temporal information when predicting a saliency map from the extracted features. Specifically, either spatial decoding and temporal aggregation are performed separately, or only one of these two processes is considered for the final prediction. The existing works are hence unable to leverage the collective spatiotemporal information, which is expected to be important to video saliency~\cite{desimone1995neural, muller1994perceptual}.

\begin{figure}[!t]
  \centering
  \includegraphics[width=\linewidth]{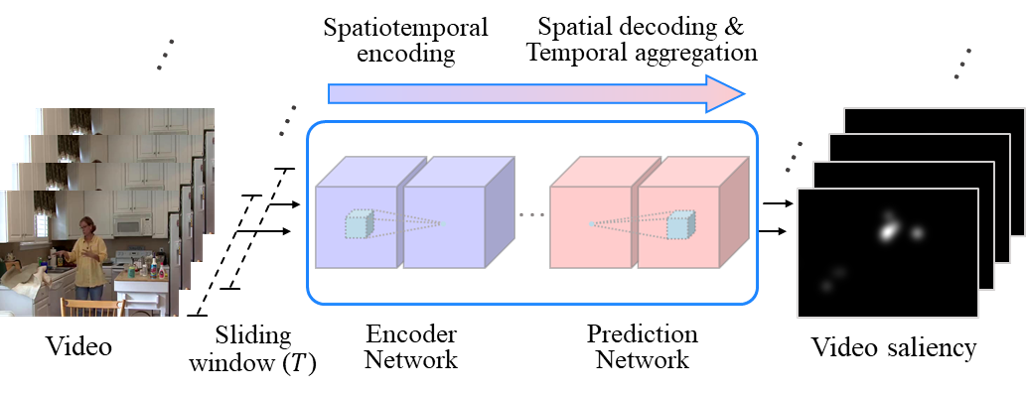}
  \caption{An illustration for the overall flow of \modelname{}. The encoder network extracts spatiotemporal features from an input clip of $T$ frames. The prediction network decodes spatially and also aggregates temporally the features to produce a single saliency map of the last input frame. This process is applied in a sliding window fashion with a window size of $T$.}
  \label{fig:overview}
\end{figure}

To this end, we propose a novel 3D fully-convolutional encoder-decoder network architecture for video saliency detection, which we call the \modelnamelong{} Network (\modelname{}). As described in Figure~\ref{fig:overview}, \modelname{} progressively reduces the temporal dimensionality within both the encoder and the decoder subnetworks, which enables it to spatially upsample the encoded features and temporally aggregate all the information as well. Similarly to other architectures designed for pixel-level tasks~\cite{badrinarayanan2015segnet, noh2015learning, ronneberger2015u}, \modelname{} compresses the spatial dimensions to extract high-level features at a low resolution, then upscales them to produce a full-resolution prediction map. On top of that, the decoder subnetwork performs temporal aggregation; we refer to it as the prediction network in our architecture since it jointly processes spatial and temporal information in a fully-convolutional way. \modelname{} predicts a single saliency map conditioned on a fixed number of previous frames, thus we apply it in a sliding-window fashion to predict a saliency map for every frame in the video.

Just as numerous 2D encoder-decoder architectures adopt VGG-16~\cite{simonyan2014very} pre-trained on ImageNet~\cite{deng2009imagenet} as their encoder network, we choose S3D~\cite{xie2018rethinking} pre-trained on the Kinetics dataset~\cite{kay2017kinetics} as the encoder network for \modelname{}. It has been shown by Xie \textit{et al.}~\cite{xie2018rethinking} that S3D is efficient and effective in extracting spatiotemporal features, and by Hara \textit{et al.}~\cite{hara2018can} that the Kinetics dataset is sufficiently large for effective transfer-learning. Therefore, we expect that the encoder network of \modelname{} can fully benefit from the successful 3D convolutional network architecture and extremely large-scale video dataset.

For the prediction network, we first place a series of transposed convolution layers and max-unpooling layers for spatial upscaling, and then we use convolution layers for temporal aggregation. The tricky part is that the max-unpooling layers cannot reuse the pooling indices or switches~\cite{zeiler2011adaptive} from the corresponding max-pooling layers since they have larger temporal receptive field than the max-unpooling layers. We introduce a new type of pooling operation, which we call \auxpname{}, that overcomes this non-trivial problem by adding extra max-poolings that can produce the properly-sized switches. \auxpname{\textit{s}} first reduce the temporal dimension of the input feature maps, and then obtain the appropriate switches for the matching max-unpooling layers. We compare \auxpname{} with two common upsampling operations, which are interpolation and transposed convolution (deconvolution), to demonstrate its effectiveness and necessity.

We comprehensively evaluate our architecture on three large-scale video saliency datasets: DHF1K~\cite{wang2018revisiting}, Hollywood2~\cite{marszalek2009actions, mathe2015actions}, and UCFSports~\cite{mathe2015actions, rodriguez2008action, soomro2014action}. Our results demonstrate that \modelname{} significantly outperforms previous state-of-the-art baselines on all three datasets. We believe that our novel architecture is effective in predicting video saliency because it jointly performs spatial decoding and temporal aggregation in a fully-convolutional way, instead of using separate recurrent units such as LSTM.

In summary, our main contributions are threefold:
\begin{itemize}[noitemsep,nolistsep,before=\vspace{4pt}, after=\vspace{4pt}]
   \item We develop a powerful end-to-end 3D fully-convolutional network for video saliency detection, comprised of an encoder network followed by a prediction network, which we name \modelname{}.
   \item We propose the novel concept of \auxpname{} which obtains switches with reduced temporal dimension so that max-unpooling layers of the prediction network can properly work.
   \item We comprehensively evaluate our proposed network on three large-scale datasets for video saliency and show the effectiveness of our joint modelling of spatial decoding and temporal aggregation.
\end{itemize}

\section{Related Work} \label{sec:related}

\begin{figure*}[ht]
  \centering
  \includegraphics[width=0.98\linewidth]{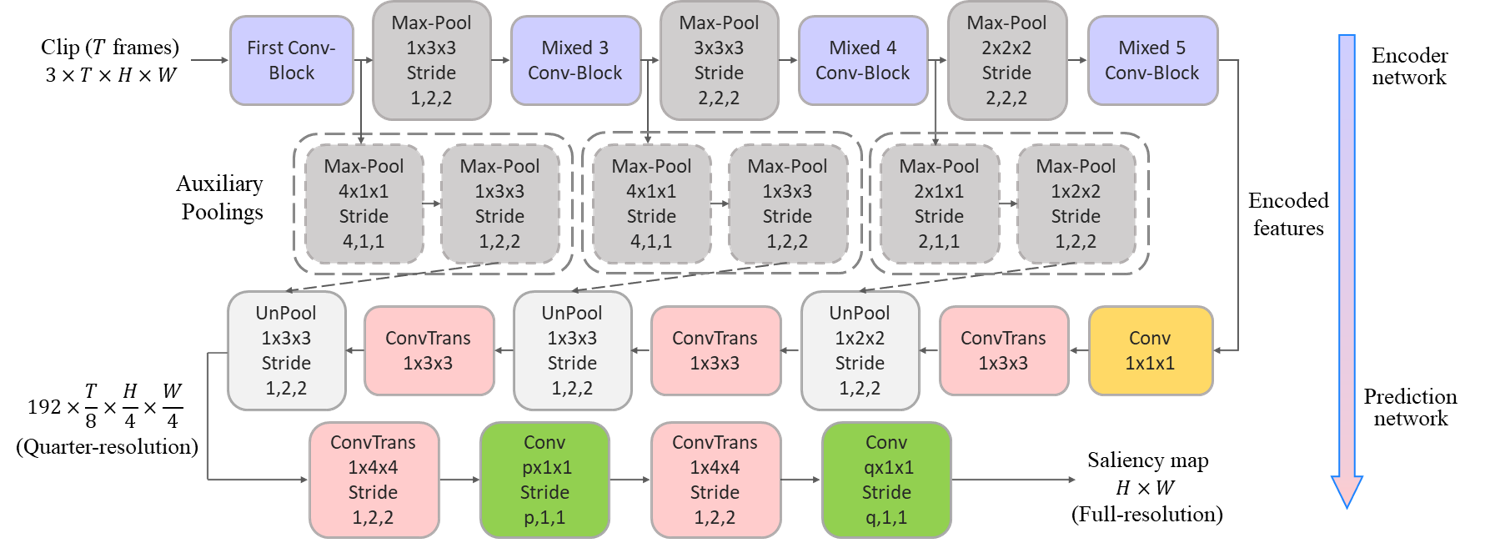}
  \caption{A detailed illustration of our proposed \modelname{} architecture. Violet boxes are convolutional operation blocks taken from the S3D~\cite{xie2018rethinking} network pre-trained on the Kinetics dataset~\cite{kay2017kinetics}. Pink boxes represent spatial decoding blocks. Green boxes are temporal convolutions that reduce the temporal dimension; within these blocks, $p$ and $q$ are set to reduce the temporal size of the output to 1. The $1\times 1\times 1$ convolutional operation in orange re-distributes the channel information of the encoded features. Because the unpooling layers operate only in spatial dimensions, switches~\cite{zeiler2011adaptive} from the pooling layers cannot be reused. \auxpname{\textit{s}} are used as extra poolings to obtain properly-sized switches for the unpooling layers. Dashed arrows represent switch transfer. Note that \auxpname{\textit{s}} are not included in the main data stream.}
  \label{fig:novel}
\end{figure*}

\textbf{Recent Video Saliency Detection Models.} Previous state-of-the-art video saliency models  rely on optical flow or LSTM to utilize temporal information. STSConvNet~\cite{bak2018spatio} adopts a two-stream architecture where temporal information from optical flow is processed independently by a temporal stream. RMDN~\cite{bazzani2016recurrent} uses spatiotemporal features extracted from C3D~\cite{tran2015learning} and then aggregates temporal information in the long term with a subsequent LSTM. OM-CNN~\cite{jiang2017predicting} first extracts spatial and temporal features from YOLO~\cite{redmon2016you} and FlowNet~\cite{dosovitskiy2015flownet} subnets, which represent objectness and motion respectively, and feed them into a two-layer LSTM. ACLNet~\cite{wang2018revisiting} implements an attention module pre-trained on SALICON~\cite{jiang2015salicon}, a large-scale dataset for image saliency, and uses the frame-wise attention mask to encourage an LSTM to better capture dynamic saliency in the long term. Comparative results of these previous models are reported in Wang \textit{et al.}~\cite{wang2018revisiting}. Image saliency detection models can also be used to predict video saliency if used in a frame-wise manner for each frame of a video. However, unsurprisingly, even state-of-the-art image saliency detection models such as SalGAN~\cite{pan2017salgan}, DVA~\cite{wang2018deep}, Deep Net~\cite{pan2016shallow}, and SALICON~\cite{huang2015salicon} are significantly outperformed by ACLNet because they does not consider any temporal information.

\textbf{Relevant 2D ConvNets.} Deep 2D ConvNets have achieved great success in diverse areas of image analysis beyond image classification for the last few years, including object detection, instance segmentation, and image saliency detection. Among such successes, VGG-16~\cite{simonyan2014very} pre-trained on ImageNet~\cite{deng2009imagenet} has played a key role as an effective feature extractor for transfer learning. Another success in 2D ConvNets has been encoder-decoder networks~\cite{badrinarayanan2015segnet, noh2015learning, ronneberger2015u}. For example, SegNet~\cite{badrinarayanan2015segnet} improves a single-stream encoder-decoder architecture by upsampling the feature maps through max-unpooling with switches from the encoder network. Switches~\cite{zeiler2011adaptive} are latent variables which record the locations of maximum activation. These variables are used by unpooling layers to partially-inverse the max-pooling operation. This method shows that max-unpooling is more suitable for decoding than other upsampling operations such as linear upsampling or even learnable upsampling method through transposed convolution, which inspires our \auxpname{}.

\textbf{Recent 3D ConvNets.} 3D ConvNets have achieved state-of-the-art results in the action recognition task. Above all, 3D ConvNets inflated from 2D ConvNets are leading the field by leveraging successful 2D network architectures as well as their parameters. Carreira and Zisserman~\cite{carreira2017quo} propose I3D, which inflates the 2D convolutional filters of Inception~\cite{szegedy2015going} to produce a 3D ConvNet with strong performance. Xie \textit{et al.}~\cite{xie2018rethinking} further explore inflated 3D ConvNets by proposing a more computationally-efficient architecture called S3D. Hara \textit{et al.}~\cite{hara2018can} experimentally show that various other inflated 3D ConvNets are also effective and predict that 3D ConvNets pre-trained on the Kinetics dataset~\cite{kay2017kinetics} can retrace the success story of 2D ConvNets, i.e. that they can be used to initialize models for many other fields of video analysis, just as VGG-16~\cite{simonyan2014very} has been applied to diverse image-based problems. We adopt S3D as the encoder network for our approach with the hope that it takes advantage of the successful architecture and the large-scale video dataset for effective transfer learning.

\section{Approach} \label{sec:approach}

\subsection{Architecture Overview} \label{subsec:overview}

The overall flow of our proposed architecture is illustrated in Figure~\ref{fig:overview}. We choose this design based on three assumptions:
\begin{inlinelist}
  \item saliency detection of any frame can be done well by only considering a fixed number of consequent past frames (we will call this number $T$ throughout this paper);
  \item given an input of $T$ frames, predicting a single saliency map for one specific time step is better than predicting maps for two or more steps at once; and
  \item there are enough number of frames in a video (specifically, the total number of frames of a video is not less than $2T-1$).
\end{inlinelist}

  The encoder network first encodes an input clip of $T$ frames spatiotemporally; this provides a deep low-resolution feature representation. Then, the following prediction network decodes the features spatially while jointly aggregating temporal information to produce a full-resolution prediction map for a single time step. We note that unlike the previous state-of-the-art models that use LSTM, our method is conditioned on a fixed number of previous frames when predicting a saliency map. The prediction network is devised to coincide with the second assumption by predicting a single saliency map corresponding to the last frame of an input clip. Frame-wise saliency maps are predicted by applying the architecture in a sliding window fashion. In other words, $S_{t}$, a saliency map at $t$, is predicted given an input clip $(I_{t-T+1},...,I_t)$ for any $t \in \{T,...,N\}$, where $I_t$ is the frame at time step $t$ and $N$ is the total number of frames in the video.
  
  The problem with this configuration is that the first $T-1$ saliency maps are not predicted. Our workaround is to reverse the chronological order of the first $T-1$ input clips. That is, $S_{t}$ for $t \in \{1,...,T-1\}$ is predicted by conditioning on $(I_{t+T-1},...,I_t)$. As a result, our architecture can predict a frame-wise saliency map for every frame as long as our third assumption that $N>=2T-1$ is satisfied. 
 
 \modelname{} has a common property with well-known image encoder-decoder networks that reduce and then upsample the spatial resolution~\cite{badrinarayanan2015segnet, noh2015learning, ronneberger2015u}. The core difference of our model comes from temporal aggregation inside the prediction network, which requires extra operations that we call \auxpname{}. The architecture of \modelname{}, along with \auxpname{}, is explained in detail in the following sections.

\subsection{Architecture specification} \label{subsec:propose}

A detailed illustration of \modelname{} is depicted in Figure~\ref{fig:novel}. An input clip is spatiotemporally encoded by 3D convolutional operation blocks of the encoder network taken from the S3D~\cite{xie2018rethinking} network pre-trained on the Kinetics dataset~\cite{kay2017kinetics}. The encoder network takes advantage of the successful 3D ConvNet architecture and the large-scale video dataset to extract rich encoded feature maps. We add a $1\times 1\times 1$ convolution after the convolutional blocks from S3D to re-distribute encoded information across the channel dimension.

Next, we describe the prediction network. We spatially upsample the encoded spatiotemporal features, leaving the time dimension alone, with a series of transpose convolutional layers and max-unpooling layers. At this point, we have only upsampled to a quarter of the original spatial resolution (quarter-resolution). Afterwards, we apply spatial transposed convolutions interspersed with temporal convolutions, which finally results in a full-resolution saliency map. The stride for these transposed convolution layers is $1\times 2\times 2$, so they double the spatial dimensions of the feature maps. The kernel sizes of the two temporal convolutions are $p\times 1\times 1$ and $q\times 1\times 1$, where $p$ and $q$ are set to 2 and $\frac{T}{16}$ respectively to aggregate all temporal information. Batch normalization~\cite{ioffe2015batch} and ReLU~\cite{nair2010rectified} come after all the convolutional operations except the last layer. After the last convolution layer, a sigmoid function is applied to produce an intensity map of saliency. A more thorough description of the architecture can be found in Supplementary material.

\subsection{\auxpnamet{}} \label{subsec:auxp}

\begin{figure}[!ht]
  \centering
  \includegraphics[width=\linewidth]{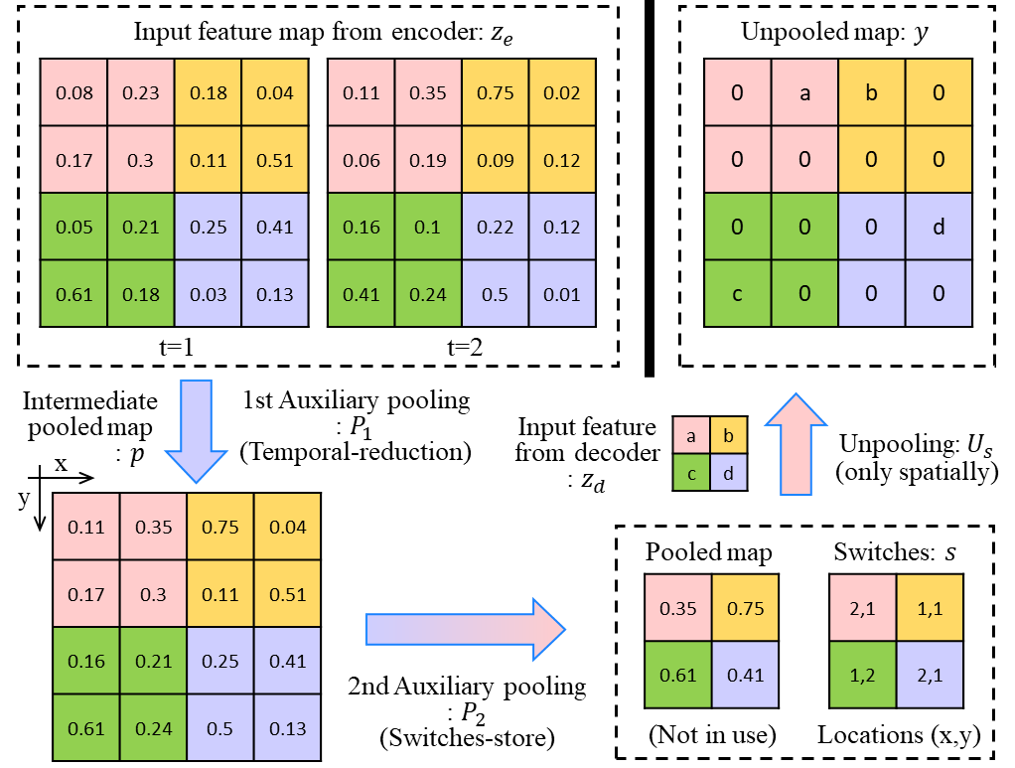}
  \caption{One example of how \auxpname{\textit{s}} work in $2\times 2\times 2$ input feature map from encoder $z_e$. The first \auxpname{} $P_1$ applies $2\times 1\times 1$ max-pooling to obtain temporally-reduced pooled map $p$. The second \auxpname{} $P_2$ applies $1\times 2\times 2$ max-pooling to store switches $s$ in the reduced temporal dimension. As a result, the corresponding unpooling layer $U_s$ with $1\times 2\times 2$ kernel can unpool the input feature map from decoder $z_d$ spatially, which produces $y$.}
  \label{fig:auxp}
\end{figure}

In our architecture, we wish to leverage the effective reconstruction ability of max-unpooling layers, which have been used in state-of-the-art pixel-level segmentation models~\cite{badrinarayanan2015segnet, noh2015learning}. However, implementing this in our architecture is non-trivial because the decoder (prediction network) never upsamples along the temporal dimension, which makes the temporal dimensions of switches~\cite{zeiler2011adaptive} from the encoder incompatible with those from the decoder. Specifically, switches of the max-unpooling layers and their corresponding max-pooling layers have different temporal sizes. In order to obtain switches with the proper sizes for the max-unpooling layers, extra processing steps are required. For each max-unpooling layers, we add two sequential extra pooling layers, which we call \auxpname{\textit{s}}. The first \auxpname{} receives the input feature map from the encoder and reduces the temporal length of the feature map. Then, the following \auxpname{}, whose kernel works only spatially, stores the proper switches for the matched unpooling layer which also only works in spatial dimension. These blocks of two sequential \auxpname{\textit{s}} make it possible for the decoder to reconstruct spatial information effectively by using the stored switches. Note that \auxpname{\textit{s}} are only used for storing switches and are not included in the main data stream. A detailed illustration of how \auxpname{\textit{s}} truly work is described in Figure~\ref{fig:auxp}. A general pooling operation $P$ takes an input feature map $z$ and produces pooled map $p$ with switches $s$ which record the location of maximum activation within the input: $[p, s] = P(z)$. The first \auxpname{} is applied to obtain the intermediate temporally-reduced pooled map $p$: $[p, \mhyphen] = P_1(z_e)$ (hyphen: variables not in use). The second \auxpname{} is applied to store switches in the reduced temporal domain: $[\mhyphen, s] = P_2(p)$. The matched unpooling operation $U_s$ unpools the input feature map from decoder only spatially using the switches $s$: $y=U_s(z_d)$. A more detailed input and output sizes can be found in Supplementary material. The necessity of \auxpname{} in \modelname{} and its variants are also further discussed in Section~\ref{subsec:aux}.

\subsection{Temporal aggregation strategy}

\begin{figure}%
\centering\begin{tabular}{c}
\hspace{1em}\includegraphics[scale=0.56]{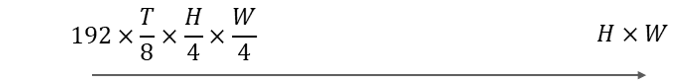}\\[1ex]
\raisebox{1.5\height}{\makebox[3em]{(a)\hspace{4.55em}}}\includegraphics[scale=0.56]{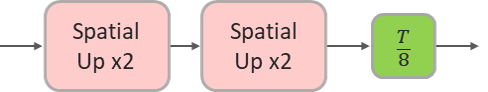}\\[1ex]
\raisebox{1.5\height}{\makebox[3em]{(b)\hspace{2.4em}}}\includegraphics[scale=0.56]{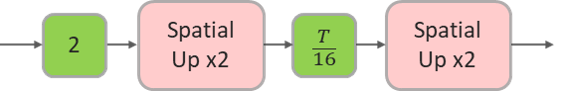}\\[1ex]
\raisebox{1.5\height}{\makebox[3em]{(c)\hspace{2.31em}}}\includegraphics[scale=0.56]{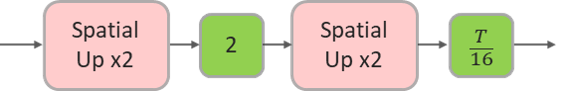}\\[0.2ex]
\end{tabular}
\caption{Different temporal aggregation strategies. Pink boxes are transposed convolutions that double each spatial dimension of the input feature maps. Green boxes are temporal convolutions that reduce the temporal dimension by a factor of the number written in each box.}
\label{fig:agg}%
\end{figure}

Temporal aggregation takes a spatiotemporally encoded feature map, whose spatial resolution is a quarter of the full video resolution, and performs the following two operations: reducing the time dimension of the input features to 1, and upscaling the spatial dimensions to full-resolution. There exist a variety of strategies that perform the required spatial upsampling and temporal reduction operations in different orders; we depict a few in Figure~\ref{fig:agg}. The first strategy, late aggregation, performs two spatial upsampling operations followed by one temporal convolutional operation that performs temporal dimension reduction. The second strategy, early two-step aggregation, performs one temporal convolution before each spatial upsampling operation. The final strategy, late two-step aggregation, performs one temporal convolution after each spatial upsampling operation. We found that late two-step aggregation performs best (see Section~\ref{subsec:on-dhf1k}), so we implemented it in \modelname{}.

\section{Evaluation} \label{sec:eval}

\subsection{Experiments setup} \label{subsec:training}
\textbf{Datasets.} We evaluate our method on three standard datasets: DHF1K~\cite{wang2018revisiting}, Hollywood2~\cite{marszalek2009actions, mathe2015actions}, and UCFSports~\cite{mathe2015actions, rodriguez2008action, soomro2014action}. These datasets and some others are compared in terms of variety, scalability, and generality by Wang \textit{et al.}~\cite{wang2018revisiting}, and we choose the DHF1K dataset as our main benchmark (i.e. we focus our analysis on this dataset) because it includes the most general and diverse scenes with various types of objects, motion, and backgrounds out of the aforementioned datasets. It consists of 1K videos with around 600K frames; 300 videos are preserved as a test set with no public ground-truth annotations of human eye fixation points. There is a public server for reporting results on the test set for fair evaluation. The Hollywood2 dataset contains 1,707 videos focusing on human actions in movie scenes, and the UCFSports dataset contains of 150 videos of human actions in sports. We believe that our selection of three datasets is sufficient to show the effectiveness and generality of our approach.

\textbf{Training/testing process.} For training \modelname{}, clips with $T$ consequent frames are randomly but densely sampled from a video. Note that this sampling scheme is valid because our model predicts each saliency map independently. Each frame is resized to $224 \times 384$. We train our network with a batch size of 40 on 600 videos from the DHF1K training set through the SGD algorithm with 0.9 momentum in an end-to-end manner. The learning rate is fixed at 0.001 for the encoder network. For the prediction network, the learning rate starts at 0.1 and decays twice by a factor of 10 when the validation loss does not decrease for a certain number of steps that depends on $T$. For \modelname{} with $T=32$, the first decaying point is at step 750, the second one is at step 950. The whole training process of 1K iterations takes less than 3 hours. Evaluation on the whole validation set takes a lot of time due to a large number of frames (60K in the validation set of the DHF1K dataset), so we uniformly sample 2K clips to approximate the validation loss. We choose Kullback-Leibler (KL) divergence as the loss function, which Jiang \textit{et al.}~\cite{jiang2015salicon} have shown to be effective for training saliency models. When testing, we apply \modelname{} in a sliding-window fashion to predict a frame-wise saliency map for every frame of all videos within the dataset. It takes around 0.06s to process each frame.

\textbf{Evaluation metrics.} Following prior work~\cite{wang2018revisiting}, we report our model's performance using the following metrics:
\begin{inlinelist}
 \item Normalized Scanpath Saliency (NSS),
 \item Linear Correlation Coefficient (CC),
 \item Similarity (SIM),
 \item Area Under the Curve by Judd (AUC-J), and
 \item Shuffled-AUC (s-AUC).
\end{inlinelist}
 NSS and CC estimate a linear correlation between the prediction and ground-truth fixation map. SIM is for computing similarity between two histograms, and AUC-J and s-AUC are variants of the well-known AUC metric. Higher scores on each metric indicate better performance.

\subsection{Evaluation on DHF1K} \label{subsec:on-dhf1k}

Since the ground-truth annotations for the test set of DHF1K~\cite{wang2018revisiting} are hidden for fair comparison, we first evaluate variants of our model on the validation set. The performance of \modelname{} with different $T$ and temporal aggregation strategies are compared in Table~\ref{table:val}. The results indicate that \modelname{} with $T=32$ and late two-step aggregation performs the best since this configuration achieves the best performance across most metrics (it has 21.2M Params and 63.2G FLOPs; more results on different $T$'s are provided in Section~\ref{subsec:dense}). We believe that late two-step aggregation performs better than early two-step aggregation because the feature maps used in spatial upscaling have a larger size in the temporal dimension. That is, late two-step aggregation performs better thanks to temporally richer feature maps. Interestingly, late aggregation performs poorly despite having the richest features, probably due to overfitting. In addition, we observe that the scores drop by 0.5 NSS (0.06 CC, 0.04 SIM, 0.015 AUC) without Kinetics pre-training for most cases. This shows the effectiveness of Kinetics pre-training. For the rest of the paper, we report the performance of \modelname{} with $T=32$, late two-step aggregation, and pre-training.

\begin{table}[!t]
\centering
\resizebox{\columnwidth}{!}{%
\begin{tabular}{|p{3.2cm}|c|c|c|c|c|} \hline
\rule{0pt}{13pt}Aggregation strategy & NSS & CC & SIM & AUC-J & s-AUC \\[3pt]\hline 
\rule{0pt}{10pt}Late-aggregation (16) & 2.555 & 0.460 & 0.340 & 0.892 & 0.712 \\
Late-aggregation (32) & 2.618 & 0.467 & 0.343 & \textbf{0.897} & 0.713 \\[0.6pt] \hline
\rule{0pt}{10pt}Early two-step (16) & 2.591 & 0.464 & 0.343 & 0.894 & 0.708\\
Early two-step (32) & 2.673 & 0.475 & 0.361 & 0.891 & 0.706 \\[0.6pt] \hline
\rule{0pt}{10pt}Late two-step (16) & 2.622 & 0.469 & 0.349 & 0.892 & 0.713 \\
\textbf{Late two-step (32)} & \textbf{2.706} & \textbf{0.481} & \textbf{0.362} & 0.894 & \textbf{0.718} \\ \hline
\end{tabular}
}
\caption{Performance comparison of \modelname{} with different $T$s (shown in parentheses) and temporal aggregation strategies on the validation set of DHF1K~\cite{wang2018revisiting}. The late two-step approach performs the best since it utilizes temporally rich features while avoiding overfitting.}
\label{table:val}
\end{table}

Next, we submitted our results to the DHF1K online benchmark~\cite{wang2018revisiting}. The performance of \modelname{} and previous state-of-the-art methods on the test set of DHF1K is reported in Table~\ref{table:test}. Our model outperforms other methods by a wide margin across all evaluation metrics. We note that ACLNet~\cite{wang2018revisiting}, the leading state-of-the-art method, is arguably better-primed for saliency detection than \modelname{}---it has a component pre-trained on an image-saliency dataset, SALICON~\cite{jiang2015salicon}, whereas we pre-train the encoder network of \modelname{} on an action recognition dataset. The higher performance of \modelname{} suggests that pre-training on a large-scale video dataset plays a significant role in performing well on other tasks in general. We also want to point out that \modelname{} has a much smaller network size (82MB v.s. 252MB). Interestingly, our AUC-J score does not increase much compared to the other metrics. This phenomenon has already been reported by Bylinskii \textit{et al.}~\cite{bylinskii2018different}, who suggest that AUC-J is less capable of discriminating between different high-performing saliency models because it is invariant to monotonic transformations.

\begin{table}[!t]
\centering
\resizebox{\columnwidth}{!}{%
\begin{tabular}{|l|c|c|c|c|c|} \hline
\backslashbox[8em]{Method}{\raisebox{-0.22\height}{Metric}} & \raisebox{-0.19\height}{\makebox[2em]{NSS}} & \raisebox{-0.19\height}{\makebox[2em]{CC}} & \raisebox{-0.19\height}{\makebox[2em]{SIM}} & \raisebox{-0.19\height}{\makebox[2.4em]{AUC-J}} & \raisebox{-0.19\height}{\makebox[2.4em]{s-AUC}} \\\hline 
\rule{0pt}{11pt}GBVS~\cite{harel2007graph} & 1.474 & 0.283 & 0.186 & 0.828 & 0.554 \\
STSConvNet~\cite{bak2018spatio} & 1.632 & 0.325 & 0.197 & 0.834 & 0.581 \\
Deep Net~\cite{pan2016shallow} & 1.775 & 0.331 & 0.201 & 0.855 & 0.592 \\ 
SALICON~\cite{jiang2015salicon} & 1.901 & 0.327 & 0.232 & 0.857 & 0.590 \\ 
OM-CNN~\cite{jiang2017predicting} & 1.911 & 0.344 & 0.256 & 0.856 & 0.583 \\ 
DVA~\cite{wang2018deep} & 2.013 & 0.358 & 0.262 & 0.860 & 0.595 \\ 
SalGAN~\cite{pan2017salgan} & 2.043 & 0.370 & 0.262 & 0.866 & 0.709 \\
ACLNet~\cite{wang2018revisiting} & 2.354 & 0.434 & 0.315 & 0.890 & 0.601 \\ 
\textbf{\modelname{}} & \textbf{2.667} & \textbf{0.470} & \textbf{0.361} & \textbf{0.895} & \textbf{0.712} \\ \hline
\end{tabular}
}
\caption{Comparison of \modelname{} with other state-of-the-art methods on the test set of DHF1K. \modelname{} significantly outperforms all the previous methods across all the evaluation metrics by a large margin.}
\label{table:test}
\end{table}

\begin{figure*}[!t]
  \adjustbox{valign=t}{\begin{minipage}[t]{0.49\linewidth}
    \includegraphics[width=1\linewidth]{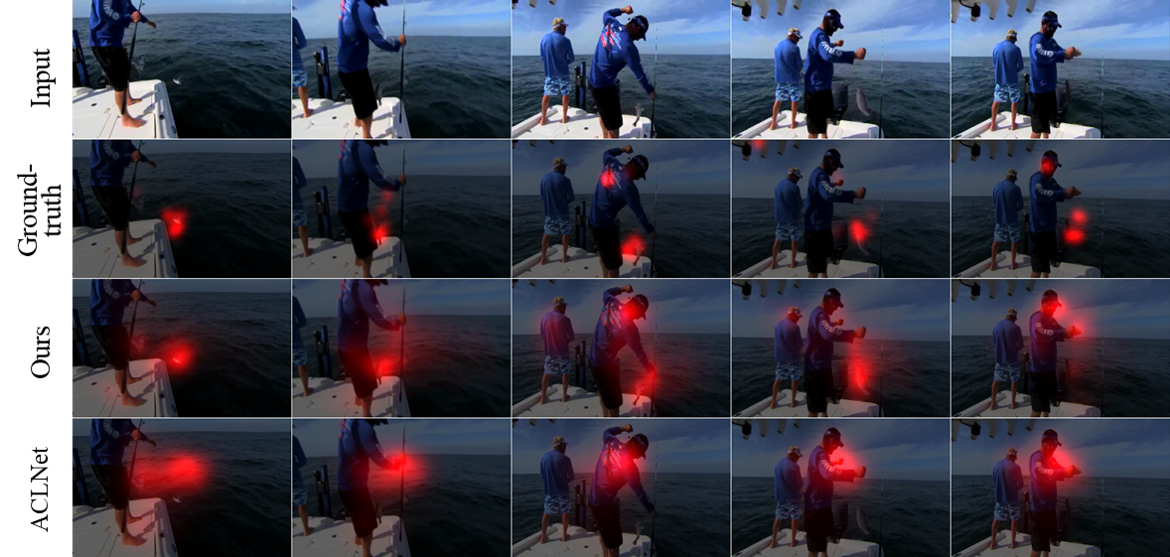}\\ \hspace*{12.35em}(a)\\[1.1ex]
    \includegraphics[width=1\linewidth]{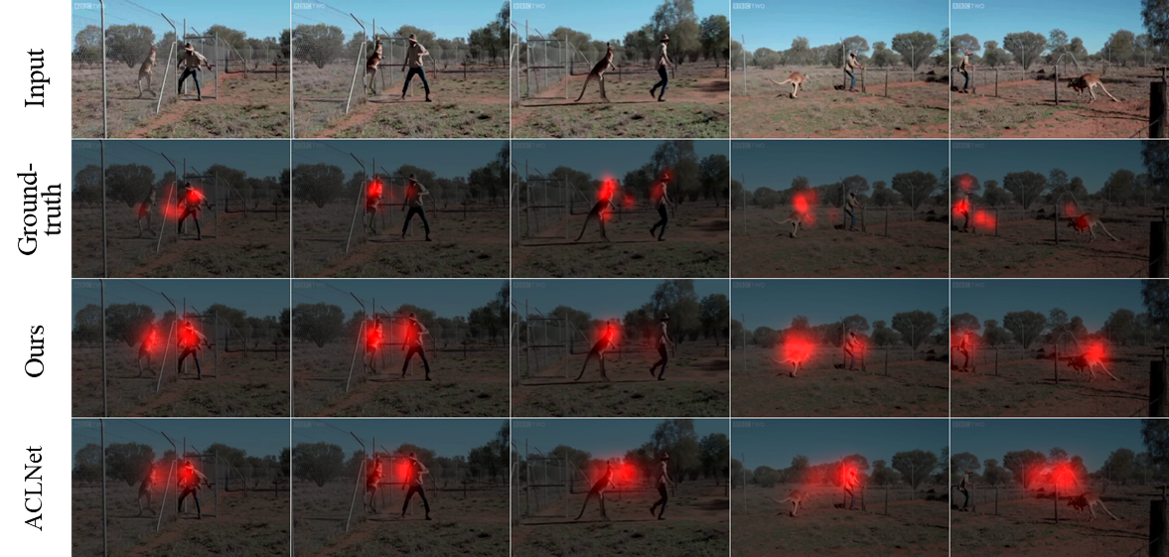}\\ \hspace*{12.35em}(b)\\
    \hspace*{2.6em}\includegraphics[width=0.85\linewidth]{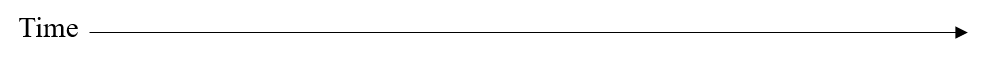}
  \end{minipage}}
  \hspace{0.03cm}\vline\hspace{0.03cm}
  \adjustbox{valign=t}{\begin{minipage}[t]{0.49\linewidth}
    \includegraphics[width=1\linewidth]{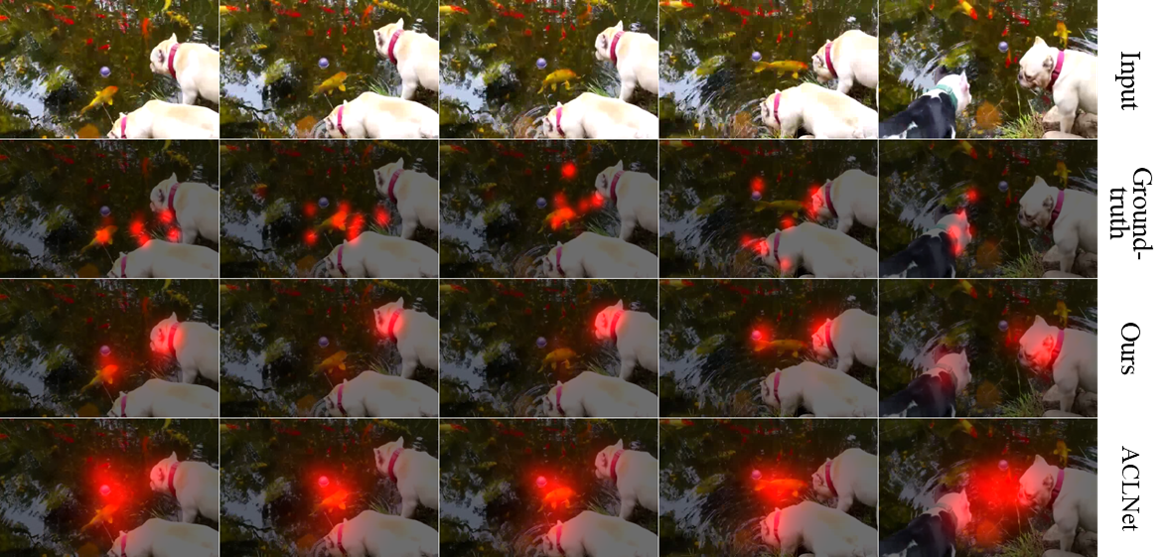}\\ \hspace*{10.8em}(c)\\[1.1ex]
    \includegraphics[width=1\linewidth]{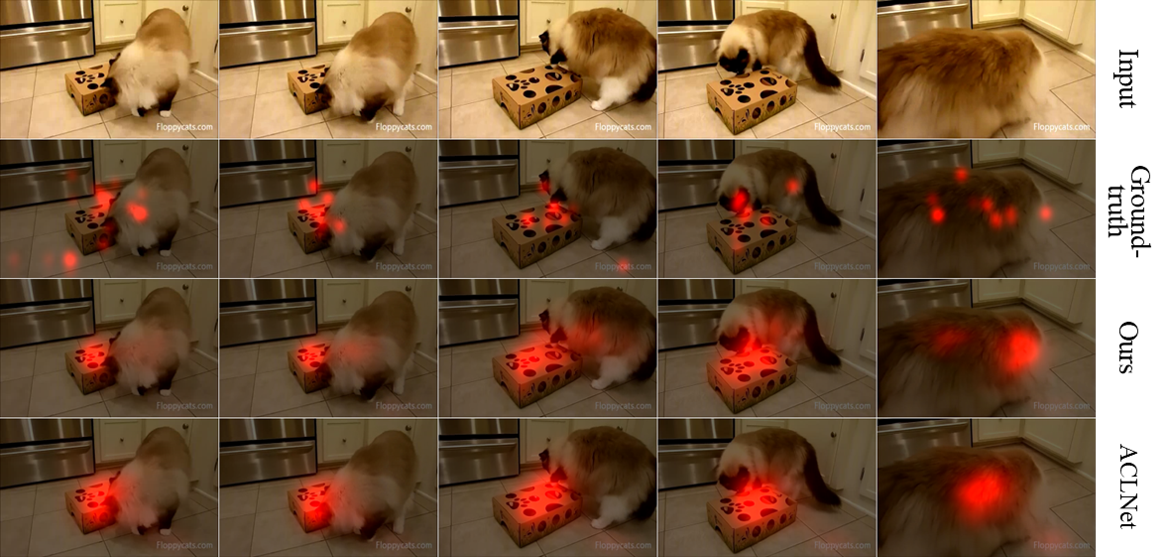}\\ \hspace*{10.8em}(d)\\
    \hspace*{1em}\includegraphics[width=0.85\linewidth]{figures/ex-t.png}
  \end{minipage}}
  \caption{Qualitative results of our \modelname{} and the main competitive model ACLNet~\cite{wang2018revisiting} on the DHF1K validation set. We observe that the differences between the two results are easily identified when the difference between NSS scores is greater than 0.5. Our method beats ACLNet by this margin on 37 videos, and ACLNet beats our method by this margin on 7 videos. We show improved results on two clips from the 37 videos ((a) and (b)), and worse results on two clips from the 7 videos ((c) and (d)). As seen in (a) and (b), \modelname{} attends to the salient moving objects very well, even when there are many background objects. In (c) and (d), it seems that the ground-truth fixation points do not represent general human gaze behavior well. For example, in (c), the fixation points flicker and jump around on different carp. In (d), only small parts of the foreground objects (the body of the cat) are fixated on. More examples are available in Supplementary material.}
  \label{fig:qual}
\end{figure*}

To perform a qualitative analysis, we compare the performance of \modelname{} to the leading state-of-the-art method, ACLNet~\cite{wang2018revisiting}, on videos from the validation set of the DHF1K dataset. We observe that we can easily recognize the differences between the results of each model when the difference of NSS scores between the two is greater than 0.5. Based on this gap, \modelname{} outperforms ACLNet on 37 out of the 100 videos in the validation set, while ACLNet outperforms \modelname{} only on 7 videos. Qualitative results of our model and ACLNet for the better and worse cases are given in Figure~\ref{fig:qual} (see Supplementary material for more examples of qualitative results). As shown in (a) and (b) in Figure~\ref{fig:qual}, \modelname{} seems highly sensitive to salient moving objects and less sensitive to background objects, which is consistent with the goal of video saliency in general. On the other hand, ACLNet seems to put more weight on spatially conspicuous objects, so sometimes it attends to distracting background objects. This makes the saliency map predicted by ACLNet a lot blurrier than ours in many cases.

We have observed that for videos where the ground-truth fixation points are scattered across a large area, our model quantitatively performs worse than ACLNet. This is because ACLNet generally predicts blurrier maps that better fit highly-scattered fixation points. However, we also find that ground-truth fixation points are unstable for these videos. For example, in (c) of Figure~\ref{fig:qual}, the fixation points do not smoothly follow the carp, but instead flicker and jump between different carp. In (d), because the foreground object is so large, fixation points tend to move around the object. Furthermore, different subjects do not fixate on the same part of a large object. In these cases, it is hard to say that the ground-truth fixation points represent general human gaze behavior well. Therefore, we strongly believe that a larger number of human subjects is needed to properly annotate videos where the fixation points are frequently scattered across a large area. We also believe that a larger and more comprehensive dataset with more diverse scenes is needed to cover general situations where the salient moving objects are not the only dominant information. More qualitative results can be found in Supplementary material.

\subsection{Performance on other datasets} \label{subsec:on-other}
We further test our model on two commonly used public datasets, which are Hollywood2~\cite{marszalek2009actions, mathe2015actions} and UCFSports~\cite{mathe2015actions, rodriguez2008action, soomro2014action}. To leverage the relatively large scale of the DHF1K dataset, we first pre-train \modelname{} on DHF1K, and then fine-tune on Hollywood2 or UCFSports. For short videos with fewer than $2T-1=63$ frames, we simply loop those videos to fit in with our method. Table~\ref{table:other} compares our model with various previous state-of-the-art approaches. \modelname{} again achieves the best performance on each dataset across most of the metrics.

\begin{table}[!t]
\centering
\resizebox{\columnwidth}{!}{%
\begin{tabular}{c|l|c|c|c|c|c} \hline
& \backslashbox[8em]{Method}{\raisebox{-0.22\height}{Metric}} & \raisebox{-0.19\height}{\makebox[2em]{NSS}} & \raisebox{-0.19\height}{\makebox[2em]{CC}} & \raisebox{-0.19\height}{\makebox[2em]{SIM}} & \raisebox{-0.19\height}{\makebox[2.4em]{AUC-J}} & \raisebox{-0.19\height}{\makebox[2.4em]{s-AUC}} \\\hline 
\rule{0pt}{10.5pt}\parbox[t]{2.4mm}{\multirow{7}{*}{\rotatebox[origin=c]{90}{Hollywood2}}} & STSConvNet~\cite{bak2018spatio} & 1.748 & 0.382 & 0.276 & 0.863 & 0.710 \\
& SALICON~\cite{jiang2015salicon} & 2.013 & 0.425 & 0.321 & 0.856 & 0.711 \\
& Deep Net~\cite{pan2016shallow} & 2.066 & 0.451 & 0.300 & 0.884 & 0.736 \\
& OM-CNN~\cite{jiang2017predicting} & 2.313 & 0.446 & 0.356 & 0.887 & 0.693 \\ 
& DVA~\cite{wang2018deep} & 2.459 & 0.482 & 0.372 & 0.886 & 0.727 \\ 
& ACLNet~\cite{wang2018revisiting} & 3.086 & 0.623 & \textbf{0.542} & 0.913 & 0.757 \\ 
& \textbf{\modelname{}} & \textbf{3.302} & \textbf{0.646} & 0.507 & \textbf{0.918} & \textbf{0.768} \\ \hline
\rule{0pt}{10.5pt}\parbox[t]{2.4mm}{\multirow{6}{*}{\rotatebox[origin=c]{90}{UCFSports}}} & GBVS~\cite{harel2007graph} & 1.818 & 0.396 & 0.274 & 0.859 & 0.697 \\
& Deep Net~\cite{pan2016shallow} & 1.903 & 0.414 & 0.282 & 0.861 & 0.719 \\
& OM-CNN~\cite{jiang2017predicting} & 2.089 & 0.405 & 0.321 & 0.870 & 0.691 \\ 
& DVA~\cite{wang2018deep} & 2.311 & 0.439 & 0.339 & 0.872 & 0.725 \\ 
& ACLNet~\cite{wang2018revisiting} & 2.567 & 0.510 & 0.406 & 0.897 & 0.744 \\ 
& \textbf{\modelname{}} & \textbf{2.920} & \textbf{0.582} & \textbf{0.469} & \textbf{0.899} & \textbf{0.752} \\ \hline
\end{tabular}
}
\caption{Comparison of \modelname{} to state-of-the-art methods on the test sets of Hollywood2 and UCFSports. High scores for our model across most of the metrics prove the effectiveness of our model.}
\label{table:other}
\end{table}

\subsection{Necessity of \auxpnamet{}} \label{subsec:aux}
As discussed earlier, \auxpname{\textit{s}} are needed for the max-unpooling layers to work in our proposed architecture. Here, we compare two possible variants of \auxpname{}. The first variant, which we call \modelname{}-tri, replaces all the max-unpooling layers with trilinear upsampling (interpolation). The second variant, which we name \modelname{}-trp, replaces the max-unpooling layers with transposed convolutions (deconvolution). Note that these two variants do not require \auxpname{\textit{s}}. Table~\ref{table:aux} compares these variants and shows that \modelname{} without \auxpname{} operations performs poorly. In other words, we discover that replacing max-unpooling layers does not work well although \modelname{}-tri and \modelname{}-trp may seem more straightforward. This proves the effectiveness and necessity of \auxpname{} in \modelname{}.

In addition, we apply our temporally-aggregating scheme to many other powerful architectures including FCN~\cite{long2015fully}, U-Net~\cite{ronneberger2015u}, Deeplab~\cite{chen2018deeplab, chen2018encoder}, which have achieved great success in dense prediction tasks. The results are reported in Supplementary material. The unsatisfying results justify our architecture with the proposed \auxpname{}.

\begin{table}[!t]
\centering
\resizebox{\columnwidth}{!}{%
\begin{tabular}{|l|c|c|c|c|c|} \hline
\backslashbox[8em]{Method}{\raisebox{-0.22\height}{Metric}} & \raisebox{-0.19\height}{\makebox[2em]{NSS}} & \raisebox{-0.19\height}{\makebox[2em]{CC}} & \raisebox{-0.19\height}{\makebox[2em]{SIM}} & \raisebox{-0.19\height}{\makebox[2.4em]{AUC-J}} & \raisebox{-0.19\height}{\makebox[2.4em]{s-AUC}} \\\hline 
\rule{0pt}{10.6pt}\modelname{}-tri & 2.452 & 0.448 & 0.337 & 0.891 & 0.702 \\
\modelname{}-trp & 2.598 & 0.470 & 0.353 & 0.894 & 0.707 \\
\modelname{} & 2.706 & 0.481 & 0.362 & 0.894 & 0.718 \\ \hline
\end{tabular}
}
\caption{Comparison of variants of \auxpname{} on the validation set of DHF1K. \modelname{}-tri and \modelname{}-trp do not utilize \auxpname{} because they replace unpooling layers with trilinear upsampling (interpolation) and transposed convolution (deconvolution), respectively. \modelname{} perform better, which demonstrates the effectiveness of \auxpname{}.}
\label{table:aux}
\end{table}

\subsection{Other observations} \label{subsec:dense}
We observe that stacking multiple transposed convolution layers with stride $1\times 1\times 1$ within each spatial decoding block in the prediction network does not boost performance. To demonstrate this, we augment \modelname{} by adding two more transposed convolutional layers to each spatial decoding block. This denser (or deeper) version approximately increases the network size by 40\%, so we expect that it would yield better performance by finely decoding spatial information. However, we found that it actually yields slightly worse performance (see Supplementary material). This might be because spatial decoding is of less importance in video saliency detection than in other tasks where more precise pixel-wise outputs are required (e.g. video segmentation). Therefore, video saliency models may not necessarily benefit from stronger spatial decoding capabilities. Otherwise, it may be due to overfitting. To better understand how this phenomenon is affected by dataset size and task formulation, we would have to test the denser \modelname{} on larger datasets and alternative tasks like video segmentation.

It is also observed that predicting multiple saliency maps all at once for each sliding window decreases the overall performance when compared to predicting a single saliency map. We believe that this is because increasing the prediction space makes it harder for the decoder (prediction network) to be optimized. It shows that our temporally-aggregating scheme is more appropriate for the video saliency detection.

Furthermore, we observe that \modelname{} with $T$ larger than 32 performs worse than when $T=32$ (see Table~\ref{table:tlarge}). These results may indicate that it is sufficient to consider a fixed number of past frames for video saliency detection. However, they could also be a result of overfitting. \modelname{} with $T$ smaller than 32 also performs worse than when $T=32$, which implies that it is necessary to consider enough number of past frames with a duration of about one second for video saliency detection. We believe that further optimization on $T$ is not necessary for this paper.

\begin{table}[!t]
\centering
\resizebox{\columnwidth}{!}{%
\begin{tabular}{|l|c|c|c|c|c|} \hline
\backslashbox[8.1em]{Method}{\raisebox{-0.22\height}{Metric}} & \raisebox{-0.19\height}{\makebox[2em]{NSS}} & \raisebox{-0.19\height}{\makebox[2em]{CC}} & \raisebox{-0.19\height}{\makebox[2em]{SIM}} & \raisebox{-0.19\height}{\makebox[2.4em]{AUC-J}} & \raisebox{-0.19\height}{\makebox[2.4em]{s-AUC}} \\\hline 
\rule{0pt}{10.6pt}\modelname{} (4) & 2.434 & 0.441 & 0.327 & 0.887 & 0.689 \\
\modelname{} (8) & 2.585 & 0.460 & 0.348 & 0.889 & 0.696 \\
\modelname{} (16) & 2.622 & 0.469 & 0.349 & 0.892 & 0.713 \\
\textbf{\modelname{} (32)} & \textbf{2.706} & \textbf{0.481} & \textbf{0.362} & \textbf{0.894} & \textbf{0.718} \\
\modelname{} (48) & 2.636 & 0.472 & 0.348 & \textbf{0.894} & 0.708 \\
\modelname{} (64) & 2.554 & 0.459 & 0.336 & 0.893 & 0.702 \\ \hline
\end{tabular}
}
\caption{Performance of \modelname{} with different $T$'s (number in bracket) on the validation set of DHF1K. The clear trend is observed. \modelname{} performs well when $T=32$.}
\label{table:tlarge}
\end{table}

\section{Conclusion} \label{sec:conclusion}

We have presented \modelname{} as a novel fully-convolutional architecture for video saliency detection. The main idea is simple but effective: spatially decoding the features extracted by the encoder while jointly aggregating all the temporal information in order to produce a single full-resolution prediction map. We also propose the new concept of \auxpname{}, which enables our architecture to leverage the benefits of max-unpooling layers for reconstruction. \modelname{} significantly outperforms previous state-of-the-art methods on major video saliency detection datasets, which demonstrates the benefits of performing spatial decoding and temporal aggregation in a fully-convolutional way, as well as the benefits of conditioning on a limited amount of past information when predicting video saliency. Finally, we comprehensively analyze \modelname{} with many variants, and show that our proposed \auxpname{} is necessary and effective.


\vspace{10pt}

\noindent \textbf{Acknowledgement.} We thank Ryan Szeto for his valuable feedback and comments. We also thank Stephan Lemmer, Mohamed El Banani, and Luowei Zhou for their discussions. This research was, in part, supported by NIST grant 60NANB17D191.

{\small
\bibliographystyle{ieee_fullname}
\bibliography{ref}
}

\end{document}